\documentclass{article}
\usepackage{spconf,amsmath,graphicx}
\usepackage{multicol,multirow,makecell}
\usepackage{subcaption}
\usepackage{adjustbox}
\usepackage{url}%
\usepackage{pgfplots}

\pgfplotsset{compat=1.16}

\title{Image-to-Graph Transformers for Chemical Structure Recognition}
\name{Sanghyun Yoo \qquad Ohyun Kwon \qquad Hoshik Lee}
\address{Samsung Advanced Institute of Technology\\
\{sam.yoo, o.kwon, hoshik.lee\}@samsung.com}
\usepackage{fancyhdr}
                    \fancypagestyle{plain}{
  \fancyhf{}% Clear header/footer
  \fancyhead{}
  \fancyfoot{}
}
\fancypagestyle{firstplain}{
  \fancyhf{}% Clear header/footer
  \fancyfoot[C]{\fontsize{8}{10} \selectfont © 20XX IEEE. Personal use of this material is permitted. Permission from IEEE must be obtained for all other uses, in any current or future media, including reprinting/republishing this material for advertising or promotional purposes, creating new collective works, for resale or redistribution to servers or lists, or reuse of any copyrighted component of this work in other works.}% Left footer
  \fancyfoot[R]{}
}
\begin{document}
%\ninept
%
\maketitle\thispagestyle{firstplain}

\begin{abstract}
For several decades, chemical knowledge has been published in written text, and there have been many attempts to make it accessible, for example, by transforming such natural language text to a structured format. Although the discovered chemical itself commonly represented in an image is the most important part, the correct recognition of the molecular structure from the image in literature still remains a hard problem since they are often abbreviated to reduce the complexity and drawn in many different styles. In this paper, we present a deep learning model to extract molecular structures from images. The proposed model is designed to transform the molecular image directly into the corresponding graph, which makes it capable of handling non-atomic symbols for abbreviations. Also, by end-to-end learning approach it can fully utilize many open image-molecule pair data from various sources, and hence it is more robust to image style variation than other tools. The experimental results show that the proposed model outperforms the existing models with 17.1 \% and 12.8 \% relative improvement for well-known benchmark datasets and large molecular images that we collected from literature, respectively.
\end{abstract}
\begin{keywords}
Optical chemical structure recognition, molecular graph, deep learning, Transformer
\end{keywords}

\section{Introduction}
\label{sec:intro}

Newly discovered chemicals are published in scientific papers such as journals and patents. Their properties or synthesis methods are described in text, which can be extracted by natural language processing and text data mining techniques. However, the chemicals themselves are mainly represented as images of graphs, in which a node and an edge indicates an atom and a bond, respectively. Such chemical graphs are often drawn to have a non-atomic symbol that represents a group of atoms or even a list of groups of atoms to reduce the complexity.

Although a tool for automatic extraction of the molecular graph structures from images would be useful for many applications, only a few ones are available. Most of them are based on the hand-crafted rules and hard to cope with many different drawing styles. Also, they mainly generate a string-based representations such as SMILES (Simplified Molecular-Input Line-Entry System) \cite{smiles}. A problem of SMILES is that it is not robust to small changes, which can result in the generation of invalid or highly different structures \cite{Nicola-2018}. Another problem is that its syntax does not permit non-atomic symbols for abbreviations.

One of the alternatives to avoid the problems of SMILES is to generate the graph structure directly. Since it is a natural representation of molecules, the transformation of the image to the graph is more intuitive than to SMILES. Also, it is easy to extend the expressiveness so that the output molecular graph can include arbitrary symbols. However, there have not been many studies to generate a graph from an image.

In this paper, we propose an end-to-end deep learning approach to recognize molecular images and generate the corresponding graphs. Our paper's main contributions are as follows: First, We propose an image encoder that consists of the ResNet and the Transformer layers. Unlike the ResNet which mainly focuses on the locality, the Transformer can capture the contextual information from a broad span in the image, and such ability enables the model to correctly recognize the bond between two atoms located far from each other in the image. Second, we design a decoder that can generate the graph structures directly from the encoded image. This decoder based on the Transformer modified to reflect the edge information outputs nodes and edges in an auto-regressive manner. Finally, we propose several training techniques suitable to our own encoder-decoder architecture. We show that our approach significantly improve the model performance.

\section{Related Work}
\label{sec:related_work}

Approaches of the chemical image recognition can be classified in two categories: the rule-based and the deep-learning-based \cite{Kohulan-2020a}.

In the past, most tools for molecular graph recognition are rule-based, which are derived from optical character recognition. 
Although they are widely used since most of them are open to public \cite{Igor-2009, Viktor-2011, Tyler-2019}, their heuristic rules often fail to cover diverse styles of images.

Due to the advances in deep learning, there recently have been many attempts to train a model that recognizes chemical images and generates molecular structures. Most of them consist of an encoder based on convolutional neural networks and a decoder that outputs SMILES \cite{Joshua-2019, Kohulan-2020b, Djork-2021, Heyley-2021} strings. Although such architecture is an easy solution, it is limited extracting only chemically valid structure since molecular structures having non-atomic symbols cannot be described in SMILES. ChemGrapher \cite{Martijn-2020} can generate a graph instead of SMILES, and hence it can be extended to handle non-atomic symbols. However, since it outputs the graph information per each image pixel, it needs images labeled pixelwise to train. This hinders the use of diverse images as the training data, although many sources of image-molecule pairs published by other organizations such as USPTO (United States Patent and Trademark Office) \cite{uspto} are available.

\section{Proposed Model}
\label{sec:proposed_model}

% In this Section, we describe our model, which consists of an image encoder and a graph decoder.

\subsection{Image Encoder}
\begin{figure*}[htbp!]
    \centering
    \includegraphics[width=0.81\textwidth]{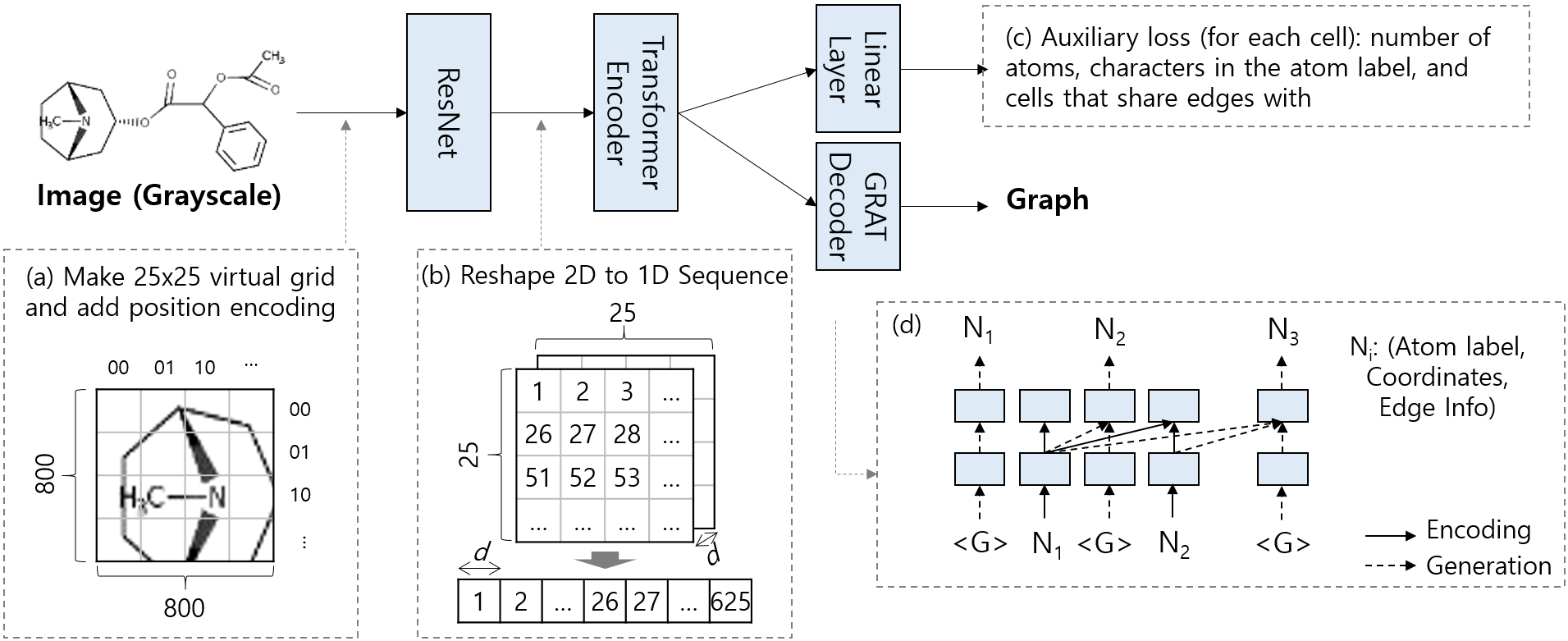}
    \caption[]%
    {{\small Encoder-decoder architecture of our model}}    
    \label{fig:enc_dec_arch}
\end{figure*}

The encoder extracts the information for generating graphs from images. The resolution of the input image of our model is fixed as 800 x 800 pixels. We virtually split the image into 25 pieces horizontally and vertically again so that there are 625 pieces each of which has 32 x 32 pixels. Then we assign the 2D position information for each piece, which is subsequently concatenated to the corresponding raw image features as in Figure \ref{fig:enc_dec_arch} (a). The output features of the ResNet-34 are reshaped to 1D sequence as shown in Figure \ref{fig:enc_dec_arch} (b). Then, they are fed into the Transformer encoder \cite{Ashish-2017} so that the relation between two atoms far from each other in the image can be captured. The final output of the encoder will be used as the input of the graph decoder explained in Section \ref{subsec:graph_decoder}.

We hope that all the atom and bond information contained in each piece of the 2D image is encoded into the corresponding step of the sequence so that our decoder can focus on small parts of the sequence when generates each node and the edges. To help the encoder gather the necessary information, we add classifiers on the Transformer encoder to predict the number of atoms, characters in the atom labels, and the IDs of pieces that share edges with for each step of the sequence, i.e., each piece in the image, as shown in Figure \ref{fig:enc_dec_arch} (c). These auxiliary losses help train the encoder in the early stages and then gradually decrease after certain steps until they reach zeros.

%Encoder는 Image로부터 Graph를 생성하는데 필요한 정보를 추출하는 역할을 수행함. ResNet을 이용하여 Image를 어느 정도 Encoding한 후에 이를 Grid 형태로 나누고 각 Grid Cell 내의 정보를 Transformer 구조를 이용하여 Vector로 변환한다. 변환된 Vector는 Grid의 맨 위쪽에서 아래쪽, 그리고 맨 왼쪽에서 오른쪽의 순서로 Sequence를 형성한다. 형성된 Sequence는 Decoder의 입력으로 들어간다.

%Transformer 구조를 활용하여 얻을 수 있는 이점은 다음과 같다. 명확히 구역을 나누어 encoding 함으로써 Decoder가 Graph를 생성할 때 집중해야 하는 부분을 찾는 부담을 줄여준다. position encoding 사용.
%Self-attention을 이용하여 인근 Cell 뿐 아니라 멀리 떨어진 Cell의 정보를 참조할 수 있다. 일반적인 이미지와 달리, 멀리 떨어져 있는 원자 사이에 결합이 존재할 수도 있기 때문이다.

%To speed up the training and help the last embedding layer of the encoder contain the necessary information, we attach, on top of each cell, an auxiliary classifier of the number of atoms, the characters in the labels of atoms, and the indices of source/destination cells from/to where the bond is.
%Encoder가 효과적으로 학습되도록 하기 위해, Guide를 마련하였다., Grid Cell 별로 생성된 vector들을 이용하여 그 안에 존재하는 원자들의 개수, 원자 label character, 다른 cell 사이의 bond 존재 유무 등을 맞추도록 하였다. 이는 Graph 생성에 필요한 정보들을 효과적으로 담도록 도와주는 역할을 한다.

\subsection{Graph Decoder}
\label{subsec:graph_decoder}
Our decoder generates the final graph by referring to the embedded sequence created by the image encoder in an auto-regressive manner; at step $t$, it first encodes the sub-graph generated at step $t$-1 and then generates a new node and its associated edges to the existing ones.

The decoder encodes the sub-graph by considering nodes as tokens of the Transformer decoder and handling edges with the self-attention, which is originally designed for handling the relation between tokens. Since the edge information (e.g., bond types) in the graph should be reflected additionally, we design our graph decoder to modulate the self-attention weights according to the edge information. To make our model learn the importance of each connection solely from the data, we use feature-wise transformation \cite{Dumoulin-2018}. In this method, the original attention weights are transformed with the scaling factor $\gamma$ and the biasing factor $\beta$, which are generated based on the edge type as the conditioning information. The final attention values, $Att(Q, K, V)$, are as follows:
\begin{gather}
	%(\gamma_{ij}, \beta_{ij}) = f_a(e_{ij}, dist_{ij})\\
	(\gamma_{ij}, \beta_{ij}) = f(e_{ij})\\
	Att(Q, K, V) = softmax \left( \frac{\Gamma \odot (QK^{T}) + B}{\sqrt{d_k}} \right) V
\end{gather}

\noindent where $e_{ij}$ indicates the edge type (in one-hot representation) between nodes $i$ and $j$, and $f$ is multi-layer perceptrons (MLPs). $\Gamma$ and $B$ are the matrices whose elements at $(i, j)$ position are $\gamma_{ij}$ and $\beta_{ij}$, respectively. The operator $\odot$ indicates the element-wise multiplication.

After the sub-graph is encoded, the graph decoder takes a special token and makes the embedding of a new node by referring to the embeddings of the sub-graph. Then, the new node and its associated edges are generated from the embedding as in Figure \ref{fig:enc_dec_arch} (d). We design a training algorithm doing such two-path decoding in parallel, explained in detail in \cite{Sanghyun-2020}. In addition to atom and bond labels, we force our graph decoder to generate coordinates in the source image. This modification helps the decoder be aware of atoms decoded so far and the atom it needs to focus on at the current step.

%Decoder는 Image Encoder가 생성한 Vector Sequence들로부터 최종 Graph를 생성하는 역할을 수행한다. Transformer Decoder를 변형하였다. 일반적으로 Transformer Decoder는 현재까지 생성한 Text와 Encoder를 기반으로 새로운 word를 생성하는 식으로 Auto-regressive하게 decoding이 진행된다. 이와 유사하게, 본 모델에서는 현재까지 생성한 sub-graph와 encoder를 기반으로 새로운 node와 edge를 생성한다. 즉, t-th step에서 decoder의 입력은 t-1 step 까지의 sub-graph 정보가 되고, 출력은 atom의 label과 t-1 step까지의 sub-graph 내 atom들과의 edge 정보가 된다.

%기본적으로 Transformer는 self-attention 구조를 이용하여 Token 간의 관계를 모델링하긴 하지만, Graph의 Edge와 같이 명시적으로 주어진 정보를 모델링하는 방법은 제공하지 않는다. 본 논문에서는 Transformer의 self-attention 계산값이 명시적으로 주어진 Edge 정보에 따라 영향을 받도록 설계함으로써 온전한 Graph 입력이 모두 사용되도록 하였다.

%Since the atom labels (e.g., C) can be repeated in a molecule, it is not easy for the decoder to remember which atom in the sub-graph is which atom in the image. To solve this problem, we exploit 2D coordinates from the image; the decoder takes the coordinates as inputs and generate the coordinates for each new atom as outputs.
%일반적으로 유기 소재의 경우 탄소(C)가 대부분을 차지하고 있으므로, 현재까지 Decoding한 C가 이미지 내의 어느 C에 해당하는지를 모델이 기억하기가 쉽지 않다. 따라서 이미지 상의 좌표 정보를 활용하도록 수정하였다. 즉, sub-graph와 각 atom의 좌표가 입력으로 들어가서 atom, edge 그리고 새 atom의 좌표가 출력으로 나온다.

\subsection{Training Data}
To train our model, we extract about 4.6 M molecules from PubChem databases and generate the corresponding images using RDKit \cite{rdkit}. As in \cite{Martijn-2020}, we modify RDKit to get the exact coordinates and characters of atom labels represented in the image, and they are used to train the encoder and the decoder. Also, to recognize a variety of superatoms, we randomly replace carbon atoms with common superatoms, e.g., CF\textsubscript{3}, NO\textsubscript{2}, t-Bu, and so on.

Although RDKit can generate various styles of images, it is not enough to cover such diverse styles of images appeared in the literature. Therefore, we additionally use 2.5 M image-molecule pairs that USPTO published. Note that the pixel-wise classification model cannot exploit these data, since there is no exact coordinate of atoms on the image. However, our end-to-end learning approach can fully utilize such data by just ignoring the losses related to the coordinates.

\section{Experiments}
\label{sec:experiments}

\subsection{Benchmark Images}

To test each model, we used four well-known benchmark images: UoB, USPTO\footnote{We made sure that images in training and test set were not overlapped.}, CLEF, and JPO. Detailed explanation of each set can be found in \cite{Kohulan-2020a}. To avoid a potential overfitting of existing tools to these well-known benchmark datasets \cite{Djork-2021}, we converted these benchmark images to PDF files and back to image files again. We insist such conversion is not too much artificial, since the recognition tools are designed to extract information from the literature which is typically published as in the PDF format. The sizes of molecules in the benchmark sets are small; average number of atoms per a molecule is around 15.8. To compare the recognition performance on the image of the large molecules, we collected 12 journal papers about organic light-emitting diodes, manually segmented 434 images of molecular structures, and generated the ground-truth molecular structures. The average number of atoms in a molecule in this dataset, called OLED, is 52.8.

%  The summary of the statistics of test sets are shown in Table \ref{tab:data_descript}.

% \begin{table}
% \centering
% \caption{\label{tab:data_descript} Statistics of five test sets used in the experiments.}
% \begin{tabular}{crrr}
% \hline   {\multirow{2}{*}{\makecell[r]{\\}}}    & {\multirow{2}{*}{\makecell[r]{Num.\\Mols}}} & {\multirow{2}{*}{\makecell[r]{Num. Valid\\SMILES}}} & {\multirow{2}{*}{\makecell[r]{Avg. Num.\\Atoms}}}  \\\\
% \hline
% UoB   & 5740     & 5740      &  13.73               \\
% USPTO & 5719     & 5704      &  31.03               \\
% CLEF  & 992      & 976       &  28.01               \\
% JPO   & 450      & 449       &  25.79            \\
% OLED  & 434      & 434       &  52.83            \\
% \hline
% \end{tabular}
% \end{table}

\subsection{Effect of Training Techniques}
We conducted several experiments to see the training techniques proposed in Section \ref{sec:proposed_model}. To see the trends quickly, we trained all models with a small portion of training data and tested on UoB benchmark images only. Although our model generates the molecular graph structure, we use the SMILES match measure (SMI) since there is no simple measure to indicate whether two graphs are exactly the same. With this measure, it is correct if the model predicts the exact same canonical SMILES as the answer. TS 1 means the ratio of the prediction whose Tanimoto similarity\footnote{The most popular similarity measures for comparing chemical structures. Two structures are usually considered similar if the value is over 0.85.} with the answer is 1.0, and Sim. means the average Tanimoto similarity.

% We also provide TS 1 measure, where the prediction is correct if its Tanimoto similarity, the most popular similarity measure for comparing chemical structures, with the answer is 1.0, as well as the average Tanimoto similarity (Sim.).

Table \ref{tab:enc_tech_results} shows the results. First we compared two base models which consist of the ResNet only. They cannot recognize images correctly regardless of coordinates prediction. When we simply put Transformer layers on the ResNet layers, the model could not be trained properly either. However, using the auxiliary loss and forcing the decoder to predict coordinates significantly boosts the accuarcy of the model.

\begin{table}
\centering
\caption{\label{tab:enc_tech_results} The effect of proposed training techniques. (O and X means 'used' and 'not used', respectively.)}
\begin{adjustbox}{width=0.9\columnwidth}
\begin{tabular}{crrrrr}
% \hline   Layers   & Aux. Loss & Coord Pred. & SMI & TS 1 & Sim. \\
\hline   {\multirow{2}{*}{\makecell[r]{\\Layers}}}    & {\multirow{2}{*}{\makecell[r]{Aux.\\Loss}}} & {\multirow{2}{*}{\makecell[r]{Coord\\Pred}}} & {\multirow{2}{*}{\makecell[r]{\\SMI}}} & {\multirow{2}{*}{\makecell[r]{\\TS 1}}} & {\multirow{2}{*}{\makecell[r]{\\Sim.}}}  \\\\
\hline
ResNet only  & -     & X      &  .000 &   .000    & .120               \\
ResNet only  & -     & O      &  .046 &   .056    &   .276               \\
ResNet + Trans  & X      & X       &  .000    &   .000    &   .000            \\
ResNet + Trans  & O      & X       &  .009    &   .010    &   .176            \\
ResNet + Trans  & O      & O       &  \textbf{.567}    &   \textbf{.615}    &   \textbf{.759}            \\
\hline
\end{tabular}
\end{adjustbox}
\end{table}

% \begin{tikzpicture}
% \begin{axis}[
% xlabel=$x$,
% ylabel=$y$,
% xmin=0, xmax=40, ymin=0, ymax=1.0,
% xtick={0, 10, 20, 30, 40},
% xticklabels={Conv only, +Coord, +Trans, +Aux.loss, +Coord},
% ytick={0, 0.1, 0.2, ..., 1.0}]
% \addplot[mark=*,blue] plot coordinates {
%     (0, 0.120)
%     (10, 0.276)
% };
% \addplot[mark=*,red] plot coordinates {
%     (20, 0.000)
%     (30, 0.176)
%     (40, 0.759)
% };
% \addlegendentry{Case 1}
% \addlegendentry{Case 2}
% \end{axis}
% \end{tikzpicture}

\subsection{Overall Performance}

Table \ref{tab:exp_results_degraded} shows the comparison of the performance among OSRA \cite{Igor-2009}, MolVec \cite{Tyler-2019}, ChemGrapher \cite{Martijn-2020}, and ours. Since we could not find the source code of ChemGrapher, we rewrite the accuracy in their paper \cite{Martijn-2020}, which was retrieved from the test on the original (i.e., not degraded) images. We trained two models: one was trained with generated images using RDKit only, and another one was trained with open image-molecule pairs, i.e., USPTO, as well as generated images.

\begin{table*}[htbp!]
\centering
\caption{\label{tab:exp_results_degraded} Performance comparison among four models tested on well-known benchmark datasets.}
\begin{adjustbox}{width=1\textwidth}
\begin{tabular}{c|ccc|ccc|ccc|ccc|ccc}
\hline & \multicolumn{3}{l|}{UOB}     & \multicolumn{3}{l|}{USPTO}   & \multicolumn{3}{l|}{CLEF}    & \multicolumn{3}{l|}{JPO}   & Total     \\ 
\cline{2-16}
\hline Model & SMI & TS 1 & Sim. & SMI & TS 1 & Sim. & SMI & TS 1 & Sim. & SMI & TS 1 & Sim. & SMI & TS 1 & Sim. \\
\hline 
OSRA v2.1.1   & .776  & .817  & .896      & .006  & .006  & .189            & .056  & .062  & .357  & .417  & .423  & .596  & .368 & .387 & .531     \\
MolVec v0.9.8 & .783  & .827 & .905      & .398  & .422  & .625     & .400  & .408  & .737   & .488  & .506  & \textbf{.758}   & .570 & .600 & .762        \\
ChemGrapher   & .706  &  - &  - &  - & - & - & - & - & - & - & - & - & - & - & -\\
Ours   & .720  & .747 & .851  &    .449   & .555 &  \textbf{.799}      & .378 & .568 & .821   & .243  & .356  & .640   & .557 & .635 & .818  \\
Ours + USPTO    & \textbf{.829}  & \textbf{.865}  & \textbf{.919}      & \textbf{.551}  & \textbf{.661}  & .792          & \textbf{.517}  & \textbf{.749}  & \textbf{.851}   & \textbf{.503}  & \textbf{.572} & .748  & \textbf{.671} & \textbf{.755} & \textbf{.852}     \\
\hline 
\end{tabular}
\end{adjustbox}
\end{table*}

Our models generally outperformed the existing models. The total accuracy over four test cases shows that our model outperformed OSRA and Molvec by 82.3 \% and 17.7 \%, respectively. Moreover, our model trained with additional USPTO data showed better performance than the one with generated data only. Our model was improved by 20.5 \% by the utilization of the large open dataset, which was possible due to our proposed image-to-graph, end-to-end learning architecture.

\begin{table}[tbp!]
\centering
\caption{\label{tab:oled_papers} Performance comparison over images extracted from actual journal papers.}
\begin{adjustbox}{width=0.75\columnwidth}
\begin{tabular}{c|rrr}
\hline Model & SMI & TS 1 & Sim. \\ \hline
OSRA v2.1.1 & .661 & .666 & .803 \\
Molvec v0.9.8 & .703 & .707 & .836 \\
Our model & .747 & .760 & .852 \\
Our model + USPTO & \textbf{.793} & \textbf{.795} & \textbf{.866} \\
\hline
\end{tabular}
\end{adjustbox}
\end{table}

Then, we tested three models on new benchmark dataset, OLED, to see their performance on the large molecules. Our model outperformed OSRA and Molvec by 20.0\% and 12.8 \% respectively as shown in Table \ref{tab:exp_results_degraded}.

\subsection{Further Analysis}

\begin{figure}[htbp]
    \centering
        \begin{subfigure}[b]{0.45\columnwidth}
            \centering
            \includegraphics[width=0.77\textwidth]{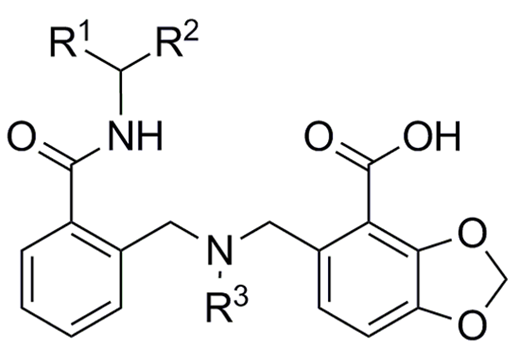}
            \caption[Network2]%
            {{\small Source}}    
            \label{fig:r1r2r3_src}
        \end{subfigure}
        \hfill
        \begin{subfigure}[b]{0.45\columnwidth}  
            \centering 
            \includegraphics[width=0.77\textwidth]{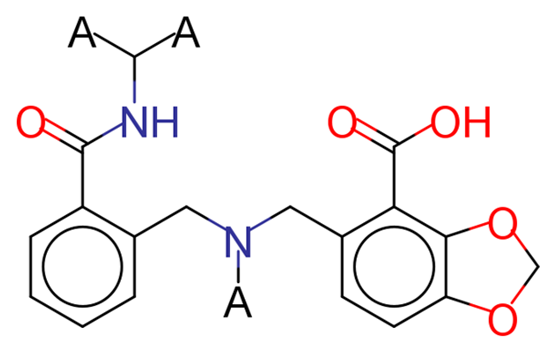}
            \caption[]%
            {{\small OSRA v2.1.1}}    
            \label{fig:r1r2r3_osra}
        \end{subfigure}
        \vskip\baselineskip
        \begin{subfigure}[b]{0.45\columnwidth}   
            \centering 
            \includegraphics[width=0.77\textwidth]{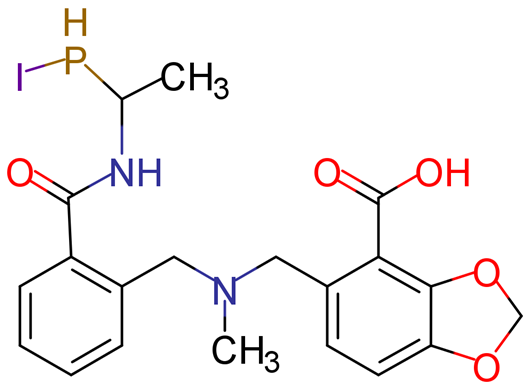}
            \caption[]%
            {{\small Molvec v0.9.8}}    
            \label{fig:r1r2r3_molvec}
        \end{subfigure}
        \hfill
        \begin{subfigure}[b]{0.45\columnwidth}   
            \centering 
            \includegraphics[width=0.77\textwidth]{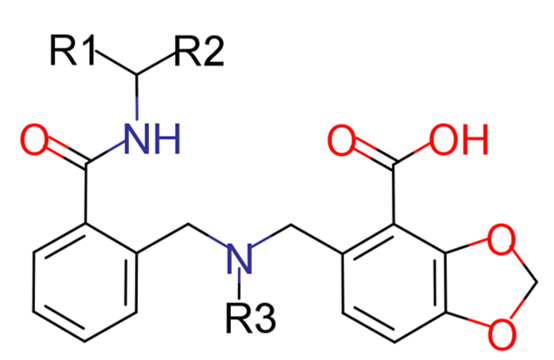}
            \caption[]%
            {{\small Our model}}    
            \label{fig:r1r2r3_i2g}
        \end{subfigure}
        \caption[ The average and standard deviation of critical parameters ]
        {\small (a) The source image of the molecule including pseudo-atom R's and recognition results of (b) OSRA, (c) Molvec, and (d) our model.} 
    \label{fig:pseudoatom_recognition_results}
\end{figure}

To show the advantage of our graph-based model, we made three models recognize the image having non-atomic symbols. Figure \ref{fig:pseudoatom_recognition_results} shows one of the source images of the molecule including pseudoatom R's and the recognition result of each model. Since OSRA can only generate SMILES that does not support non-atomic symbols, it recognized R1, R2, and R3 as A (any-atom); it is not possible to distinguish them. Molvec could not recognize them properly at all. However, our model recognized R1, R2, and R3 exactly, and hence it can substitute them with the intended sub-structures by post-processing.

\begin{figure}[htbp]
    \centering
    \begin{subfigure}[t]{0.95\columnwidth}
    \centering
        \includegraphics[width=1.0\textwidth]{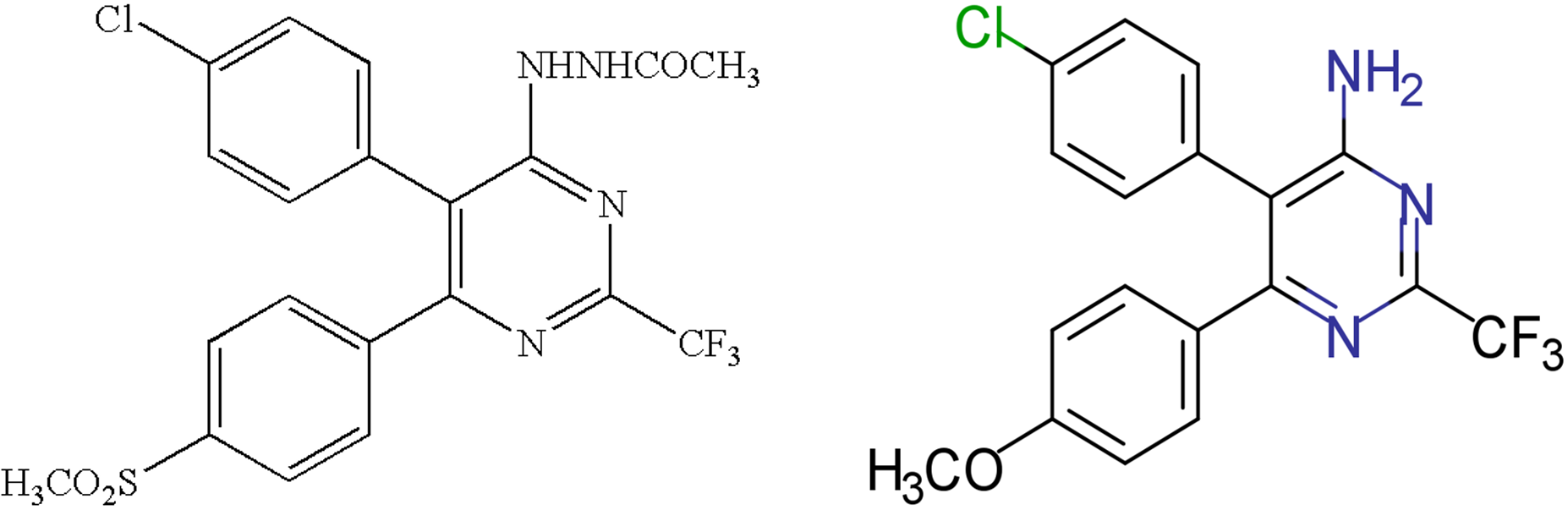}
        \label{fig:failed_img_uspto}
        \subcaption{}
    \end{subfigure}
    \begin{subfigure}[t]{0.9\columnwidth}
    \centering
        \includegraphics[width=0.55\textwidth]{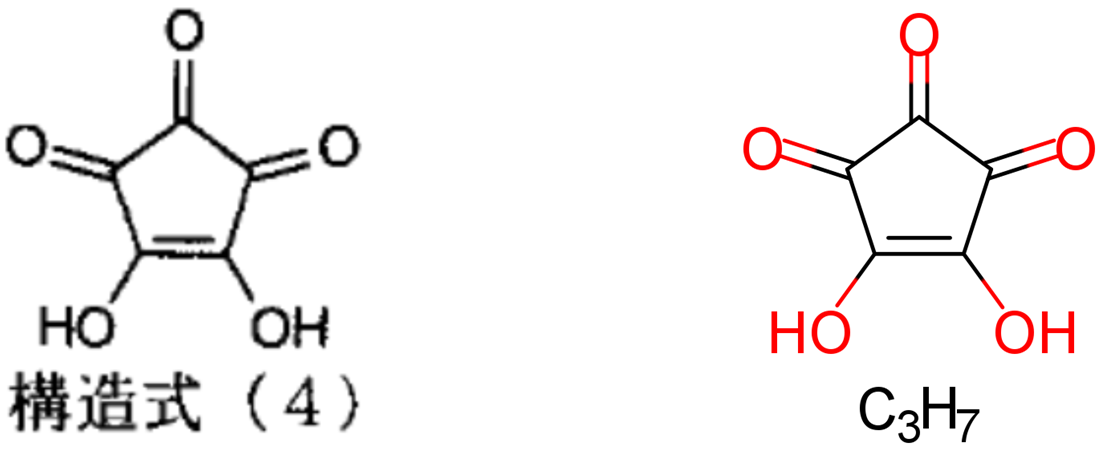}
        \label{fig:failed_img_jpo}
        \subcaption{}
    \end{subfigure}
    \caption{Examples of images that our model fails to recognize: source images (left) and the recognized results (right).}
    \label{fig:failed_img}
\end{figure}

We analyzed the recognition results of USPTO, CLEF, and JPO, since our model showed poor performance on these datasets. One of the main reasons is that there were many superatoms which our model could not handle with, e.g., NHNHCOCH\textsubscript{3} or H\textsubscript{3}CO\textsubscript{2}S in Figure \ref{fig:failed_img} (a). Although generating more superatoms in training data can be one of solutions, the ultimate solution is to recognize symbols by character-level. We plan to modify the atom classifier to character-level decoder as future work.

The other reason is that, as in Figure \ref{fig:failed_img} (b), our model tried to recognize captions as atoms. This issue was worse in JPO cases than others. Such behaviors might be reduced by augmenting training images to contain arbitrary captions so that our model could ignore them, or by designing an image segmentation model.

\section{Conclusion}
\label{sec:conclusion}

In this paper, a deep learning approach to recognize the molecular images and generate the graph structures directly even when there are non-atomic symbols for abbreviations was suggested. We also provided novel training techniques that significantly boost the accuracy of the model. Through extensive experiments over several benchmark datasets, we showed that the proposed model outperformed existing open source libraries. As future work, we plan to extend the model to generate atom labels at character-level, so that it can recognize any arbitrary symbols.
%본 논문에서는 Chemical Image를 인식하여 Graph 구조를 생성하는 모델을 제안하였다. 다양한 Superatom들의 인식이 가능하고, End-to-End 방식으로 학습이 되므로 이미지-mol pair 데이터를 활용할 수 있다는 장점이 있다. 대규모 benchmark 데이터와, 실제 논문에서 추출한 이미지 데이터를 이용하여 기존 모델들과 성능 비교를 수행했고, 성능이 월등함을 알 수 있었다.

\vfill\pagebreak

\bibliographystyle{IEEEbib}
\bibliography{arxiv}

\end{document}